\pdfoutput=1

\documentclass[11pt]{article}

\usepackage[]{acl}

\usepackage{times}
\usepackage{latexsym}

\usepackage[T1]{fontenc}

\usepackage[utf8]{inputenc}
\usepackage{microtype}
\usepackage{arydshln}

\usepackage{inconsolata}
\usepackage{url}
\usepackage{color}
\usepackage{multirow}
\usepackage{booktabs}
\usepackage{amsmath}

\usepackage{graphicx}

\title{Enhancing Input-Label Mapping in In-Context Learning with\\ Contrastive Decoding}

\author{%
  Keqin Peng$^{1}$,
  Liang Ding$^{2}$\thanks{~~Corresponding Authors.},
  Yuanxin Ouyang$^{1}$,
  Meng Fang$^{3}$,
  Yancheng Yuan$^{4}$,
  Dacheng Tao$^{5}$\\
  $^{1}$Beihang University $^{2}$The University of Sydney $^{3}$University of Liverpool\\
  $^{4}$The Hong Kong Polytechnic University $^{5}$Nanyang Technological University\\
  \texttt{keqin.peng@buaa.edu.cn},
  \texttt{liangding.liam@gmail.com}}

\begin{document}
\maketitle
\begin{abstract}
Large language models (LLMs) excel at a range of tasks through in-context learning (ICL), where only a few task examples guide their predictions. However, prior research highlights that LLMs often overlook input-label mapping information in ICL, relying more on their pre-trained knowledge. To address this issue, we introduce In-Context Contrastive Decoding (ICCD), a novel method that emphasizes input-label mapping by contrasting the output distributions between positive and negative in-context examples. Experiments on 7 natural language understanding (NLU) tasks show that our ICCD method brings consistent and significant improvement (up to +1.8 improvement on average) upon 6 different scales of LLMs without requiring additional training. Our approach is versatile, enhancing performance with various demonstration selection methods, demonstrating its broad applicability and effectiveness. The code and scripts are released at~\url{https://github.com/Romainpkq/CD_ICL}.

\end{abstract}

\section{Introduction}
In-context learning (ICL,~\citealp[]{DBLP:conf/nips/BrownMRSKDNSSAA20}) is one of the most remarkable emergent capabilities of large language models (LLMs,~\citealp[]{achiam2023gpt,dubey2024llama}). By leveraging just a few carefully selected input-output examples, ICL enables models to adapt to new tasks without parameter updating~\cite{dong2022survey,peng-etal-2024-revisiting}. This approach has proven highly effective in unlocking the advanced capabilities of LLMs and has become a standard technique for tackling a spectrum of tasks, like translation, coding, and reasoning~\cite{peng2023towards,wang2025leveraging,wibisono2024context}.

Previous studies~\cite{pan-etal-2023-context,wei2023larger} have identified two critical factors for successful ICL: \textit{task recognition (TR)}, which involves identifying the task from the demonstrations and utilizing prior knowledge to make predictions, and \textit{task learning (TL)}, which focuses on directly learning the input-label mappings from the demonstrations. However, ICL faces challenges in overcoming the biases introduced by pretraining~\cite{kossen2024context}, and LLMs tend to underutilize input-label mapping information~\cite{min-etal-2022-rethinking}. For example, in tasks like SST-2~\cite{socher2013recursive}, the model may default to using its internal knowledge rather than learning the specific input-label mappings provided in the context.

To address this issue, we propose a simple yet effective method called \textit{in-context contrastive decoding} (ICCD). Our method is inspired by the contrastive decoding technique~\cite{li-etal-2023-contrastive,sennrich2024mitigating,kim2024instructive,zhong-etal-2024-rose,wang2024mathbb}, which increases the probability of the desired output by suppressing undesired outputs, and our ICCD enhances the model’s attention to input-label mapping during generation. Specifically, we construct negative in-context examples by altering the inputs of the demonstrations, creating incorrect input-label mappings while keeping the labels unchanged. By comparing the output distributions between positive and negative examples, ICCD effectively emphasizes the correct input-label mappings, integrating this information into the original ICL process. Notably, our method works with any pretrained LLMs without requiring additional training.

Experimental results across seven natural language understanding tasks demonstrate that our ICCD strategy consistently and significantly improves performance upon several advanced LLMs, e.g., Llama-3.1, Llama-3.2, and Qwen2, across various datasets and model scales. Moreover, we show that ICCD can be seamlessly integrated with different demonstration selection methods, showcasing its robustness and universal applicability.

\section{Methodology}
\subsection{Background} 
Given an input query $x$, the probability of generating the target $y$ using a casual LLM $M$ parameterized by $\theta$ can be formulated as follows:
\begin{equation}
    y \sim p_\theta(y \mid \boldsymbol{c}, \mathcal{T}(x)),
\end{equation}
where $\mathcal{T(\cdot)}$ is the template used to wrap up inputs and $c = \mathcal{T}(x_1),\cdots, \mathcal{T}(x_k)$ is the context string concatenating $k$ in-context examples, $p_\theta(y \mid \boldsymbol{c}, \mathcal{T}(x))= \operatorname{softmax}[\ \operatorname{logit}_\theta(y \mid \boldsymbol{c}, \mathcal{T}(x))]$ is the probability for the predicted token. For obtaining the desired $y$, the regular decoding method is to choose the token with the highest probability (\textit{i.e.}, greedy decoding) or sampling from its distribution (\textit{e.g.}, top-k decoding).

Here, we can observe that there are two kinds of knowledge contributing to model prediction, models' prior knowledge and input-label mapping information in in-context learning. However, LLMs usually prioritize prior knowledge over input-label mapping information~\cite{kossen2024context}, leading to ICL's struggle to fully overcome prediction preferences acquired from pre-training.

\subsection{In-Context Contrastive Decoding} 
\label{subsec:cd}
To mitigate the issue above, we construct negative in-context examples to factor out the input-label mapping from the models' original output distribution contrastively. Specifically, in addition to the origin in-context examples $\boldsymbol{c}$, we construct negative in-context examples $\boldsymbol{c}^-$ with incorrect input-label mapping. We then subtract the negative output $\boldsymbol{\mathrm{z}}_{t}^{-}$ from the positive output $\boldsymbol{\mathrm{z}}_t$ to isolate the knowledge of input-label mapping. Finally, we integrate this knowledge with the original in-context learning to reinforce the importance of input-label mapping:
\begin{equation}
\label{eq:log}
y_t \sim \mathrm{softmax}(\boldsymbol{\mathrm{z}}_t + \alpha (\boldsymbol{\mathrm{z}}_{t} - \boldsymbol{\mathrm{z}}_{t}^{-}))\text{,}
\end{equation}
where $\alpha$ is a hyperparameter that governs the importance of input-label mapping information. Equivalently,
\begin{align}
    y_t &\sim \tilde p_{\theta}(y | \boldsymbol{c},\boldsymbol{c}^-,\mathcal{T}(x)) \\
    &\propto p_{\theta}(y | \boldsymbol{c},\mathcal{T}(x))\left( \frac{ p_{\theta}(y | \boldsymbol{c},\mathcal{T}(x)}{ p_{\theta}(y | \boldsymbol{c}^{-},\mathcal{T}(x))}\right)^{\alpha}\text{.}
\end{align}

\paragraph{Construction of $\boldsymbol{c}^-$.} The negative in-context examples $\boldsymbol{c}^-$ is the key to the success of the in-context contrastive decoding method (ICCD). Considering the label bias~\cite{zhao2021calibrate} of in-context learning, directly altering the labels of demonstrations may introduce a completely different label bias, potentially distorting the input-label mapping information. Hence, we adjust the inputs instead of the labels to change input-label mapping information. Specifically, for each demonstration ${(x_i, y_i)}$, we first randomly select a different label $y_j(y_j \neq y_i)$ from the label space. Then we randomly choose an input $x_j$ whose label is $y_j$ from the demonstrations pool to construct the negative demonstration ${(x_j, y_i)}$. We compare the effect of different $\boldsymbol{c}^-$ in Section~\ref{subsect:analysis}.

\section{Experimental Setup}
\paragraph{Models and Baselines.} We perform experiments across different sizes of models, including Llama-series: Llama3.2-1B (1B), Llama3.2-3B (3B) and Llama3.1-8B (8B) ~\cite{dubey2024llama} and Qwen2 series: Qwen2-0.5B (0.5B), Qwen2-1.5B (1.5B) and Qwen2-7B (7B)~\cite{yang2024qwen2technicalreport}, which are all widely-used decoder-only dense LMs. We also conduct experiments on extensive alignment models, e.g., Llama3.2-1B-Instruct, Llama3.2-3B-Instruct, and Llama3.1-8B-Instruct~\cite{dubey2024llama} to verify the generalizability of our approach. For the baseline, we use the regular decoding methods following prior work~\cite{shi-etal-2024-trusting,zhao-etal-2024-enhancing}. 

\paragraph{Demonstration Selection methods.} To verify that our method is complementary to different demonstration selection methods, we mainly consider three different demonstration selection methods that do not require additional training.
\begin{itemize}
 \item \textbf{Random} baseline randomly select in context examples for each testing sample.
  \item \textbf{BM25}~\cite{robertson2009probabilistic} baseline uses BM25 to calculate the word-overlap similarity between samples and test input and select the high-similarity samples as demonstrations.
  \item \textbf{TopK}~\cite{liu2022makes} baseline uses the nearest neighbors of a given test sample as the corresponding in-context examples.
\end{itemize}

\paragraph{Datasets and Metrics.} We conduct a systematic study across 7 NLU tasks, including binary, multi-class classification tasks (SST-2, SST-5~\cite{socher-etal-2013-recursive}, CR, Subj~\cite{wang-etal-2018-glue}) and natural language inference tasks: MNLI~\cite{williams-etal-2018-broad} and QNLI~\cite{wang-etal-2018-glue}. We will report the accuracy to show the performance.

\paragraph{Experimental Details.} Our method introduces a hyperparameter $\alpha$ to control the input-label mapping information. For simplicity, we set $\alpha=1$ for all models and settings. We ran all experiments 3 times with different random seeds and reported the average accuracies. We use 16-shot ICL for all models. Without a special statement, we report the results of the random selection method.

\section{Main Results}
\begin{figure}[t!]
    \centering
    \includegraphics[width=1.0\columnwidth]{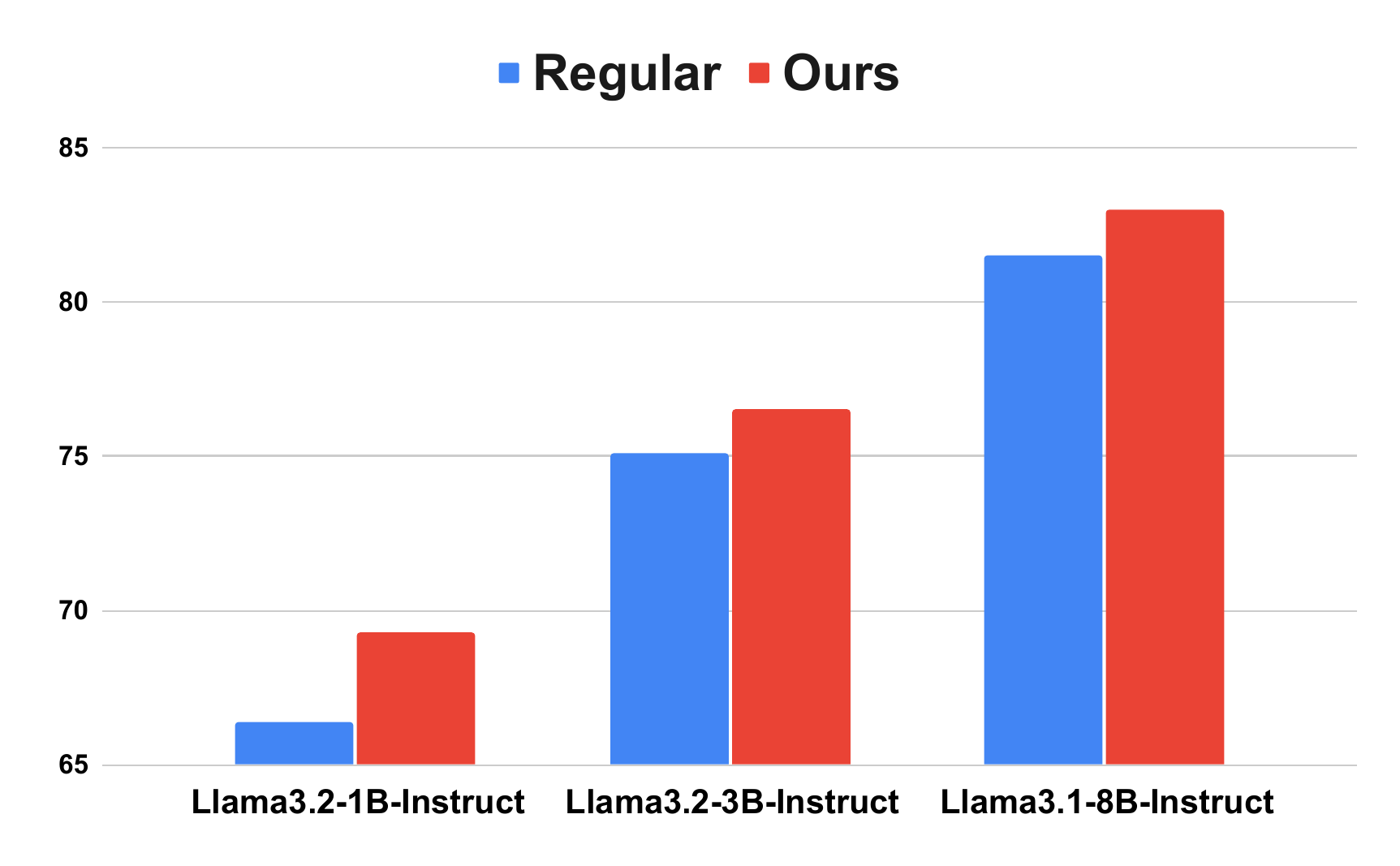}
    \caption{\textbf{Performance with different chat models.}}
    \label{fig:ins}
\end{figure}

 We demonstrate the effectiveness of our method in 7 NLU tasks described in the Datasets and Metrics section. We summarize the results in Table~\ref{tab:ret}, Table~\ref{tab:task_model}, Tabel~\ref{tab:mul_class}, and Figure~\ref{fig:ins}. Based on the results, we can find that: 
\paragraph{Our method brings gain across different tasks and model scales.} Results on Table~\ref{tab:task_model} show that our method can achieve consistently better performance across the majority of tasks under different model scales than the regular decoding method. Specifically, our method brings over 1.0 improvements (in accuracy) in all Llama-series models and Qwen2-series models. It's worth highlighting that ICCD brings +2.3 gains on average in the Qwen2-1.5B model. 
Furthermore, it is noteworthy that our approach can achieve more significant improvements in challenging tasks with the increase of model scale, such as QNLI and MNLI tasks, respectively bringing 5.1\% (1.4\%) and 1.8\% (1.2\%) gains compared to regular decoding in Llama3.1-8B (Qwen2-7B), demonstrating the effectiveness and universality of our method. 

\paragraph{Our method consistently improves the performance with different in-context examples selection methods.} Table~\ref{tab:ret} lists the average performance and standard deviation of different models with different demonstration selection methods. Clearly, our method can achieve better and stable performance with different demonstration selection methods. When the model scale increases, our method can achieve more improvement gains compared to the regular decoding method, +0.5 and +1.1 with BM25 method under Llama3.2-3B and Llama3.1-8B, respectively. These results prove that ICCD can be complementary with different demonstration selection methods.

\paragraph{Our method works for aligned chat models.} To verify the effectiveness of our method for the chat LLMs, we conducted experiments on different instruction-tuned and RLHF-tuned LLMs. Figure~\ref{fig:ins} show that our method can achieve consistent improvement in different chat models, demonstrating that our method also works for instruction-tuned and safety-enhanced models.

\paragraph{Our method works for a larger number of target classes.} To verify the effectiveness of our method for a larger number of target classes, we conducted experiments on datasets TREC (6 target classes) and Dbpedia (14 target classes) with the random selection method. Results on Table~\ref{tab:mul_class} show that our method can achieve remarkable improvement, demonstrating the effectiveness of our method in larger target classes.

\begin{table}
\centering
\scalebox{0.68}{
\begin{tabular}{cccccccc} 
\toprule
\multirow{2}{*}{\textbf{Model}}       & \multirow{2}{*}{\textbf{Decoding}} & \multicolumn{2}{c}{\textbf{Random}}            & \multicolumn{2}{c}{\textbf{BM25}}              & \multicolumn{2}{c}{\textbf{TopK}}               \\ 
\cmidrule(lr){3-4}\cmidrule(lr){5-6}\cmidrule(r){7-8}
                                      &                                    & \textbf{avg.}                  & \textbf{std.} & \textbf{avg.}                  & \textbf{std.} & \textbf{avg.}                  & \textbf{std.}  \\ 
\midrule
\multirow{2}{*}{\textbf{Llama3.2-1B}} & \textbf{Regular}                   & 66.1                           & -             & 72.5                           & -             & 73.6 & -              \\
                                      & \textbf{Ours}                      & \textcolor{red}{\textbf{68.3}} & 0.19          & \textbf{\textcolor{red}{72.9}} & 0.11          & \textcolor{green}{\textbf{73.4}}                           & 0.17           \\ 
\midrule
\multirow{2}{*}{\textbf{Llama3.2-3B}} & \textbf{Regular}                   & 72.9                           & -             & 76.6                           & -             & 76.7                           & -              \\
                                      & \textbf{Ours}                      & \textbf{\textcolor{red}{74.6}} & 0.47          & \textbf{\textcolor{red}{77.1}} & 0.28          & \textbf{\textcolor{red}{76.9}} & 0.19           \\ 
\midrule
\multirow{2}{*}{\textbf{Llama3.1-8B}} & \textbf{Regular}                   & 77.6                           & -             & 79.7                           & -             & 80.2                           & -              \\
                                      & \textbf{Ours}                      & \textbf{\textcolor{red}{79.4}} & 0.19          & \textbf{\textcolor{red}{80.8}} & 0.15          & \textbf{\textcolor{red}{80.9}} & 0.05           \\
\bottomrule
\end{tabular}
}
\caption{\textbf{Average performance and standard deviation of 7 Natural Language Understanding (NLU) tasks with different in-context example selection methods.} \textbf{\textcolor{red}{Red}} results indicate that our method brings improvement over the regular decoding, while \textbf{\textcolor{green}{Green}} results denote no improvement.}
\label{tab:ret}
\end{table}

\begin{table*}
\centering
\scalebox{0.95}{
\begin{tabular}{cccccccccc} 
\toprule
\textbf{Model}                        & \textbf{Decoding} & \textbf{SST2}                    & \textbf{CR}                    & \textbf{SST5}                    & \textbf{Subj}                  & \textbf{QNLI}                  & \textbf{MNLI}                    & \textbf{AG\_NEWS}                & \textbf{\textit{Avg.}}                 \\ 
\midrule
\multirow{2}{*}{\textbf{Llama3.2-1B}} & Regular           & 89.8                             & 83.0                           & 43.7                             & 72.8                           & 53.5                           & 36.6                             & 83.3                             & 66.1                                   \\
                                      & Ours              & \textbf{\textcolor{red}{91.1}}   & \textbf{\textcolor{red}{83.7}} & \textbf{\textcolor{green}{43.3}} & \textbf{\textcolor{red}{83.0}} & \textbf{\textcolor{red}{53.8}} & \textbf{\textcolor{red}{39.2}}   & \textbf{\textcolor{red}{84.1}}   & \textbf{\textcolor{red}{68.3 (+2.1)}}  \\ 
\hdashline
\multirow{2}{*}{\textbf{Llama3.2-3B}} & Regular           & 93.7                             & 87.2                           & 46.2                             & 86.0                           & 54.2                           & 56.9                             & 86.4                             & 72.9                                   \\
                                      & Ours              & \textbf{\textcolor{red}{94.0}}   & \textbf{\textcolor{red}{88.1}} & \textbf{\textcolor{red}{46.5}}   & \textbf{\textcolor{red}{92.1}} & \textbf{\textcolor{red}{57.2}} & \textbf{\textcolor{red}{57.0}}   & \textbf{\textcolor{red}{86.9}}   & \textbf{\textcolor{red}{74.6 (+1.7)}}  \\ 
\hdashline
\multirow{2}{*}{\textbf{Llama3.1-8B}} & Regular           & 96.7                             & 92.3                           & 48.0                             & 94.0                           & 60.3                           & 65.3                             & 86.7                             & 77.6                                   \\
                                      & Ours              & \textbf{\textcolor{green}{96.5}} & \textcolor{red}{\textbf{93.2}} & \textcolor{red}{\textbf{49.3}}   & \textcolor{red}{\textbf{96.1}} & \textcolor{red}{\textbf{65.4}} & \textcolor{red}{\textbf{67.5}}   & \textcolor{red}{\textbf{87.6}}   & \textcolor{red}{\textbf{79.4 (+1.8)}}  \\ 
\midrule
\multirow{2}{*}{\textbf{Qwen2-0.5B}}  & Regular           & 87.9                             & 89.4                           & 34.5                             & 62.2                           & 52.5                           & 47.6                             & 78.1                             & 64.6                                   \\
                                      & Ours              & \textcolor{red}{\textbf{89.2}}   & \textcolor{red}{\textbf{89.6}} & \textcolor{green}{\textbf{33.9}}   & \textcolor{red}{\textbf{68.1}} & \textcolor{red}{\textbf{53.2}} & \textbf{\textcolor{red}{47.6}} & \textbf{\textcolor{red}{78.7}} & \textcolor{red}{\textbf{65.8 (+1.2)}}  \\ 
\hdashline
\multirow{2}{*}{\textbf{Qwen2-1.5B}}  & Regular           & 95.2                             & 91.0                           & 49.0                             & 72.3                           & 60.2                           & 61.8                             & 76.7                             & 72.3                                   \\
                                      & Ours              & \textbf{\textcolor{green}{95.1}} & \textbf{\textcolor{red}{91.3}} & \textbf{\textcolor{green}{48.3}}   & \textbf{\textcolor{red}{81.5}} & \textbf{\textcolor{red}{61.8}} & \textbf{\textcolor{red}{65.2}}   & \textbf{\textcolor{red}{79.1}}   & \textbf{\textcolor{red}{74.6 (+2.3)}}  \\ 
\hdashline
\multirow{2}{*}{\textbf{Qwen2-7B}}    & Regular           & 96.0                             & 91.5                           & 51.9                             & 82.3                           & 71.4                           & 78.7                             & 83.8                             & 79.4                                   \\
                                      & Ours              & \textcolor{red}{\textbf{96.3}}   & \textcolor{red}{\textbf{91.7}} & \textcolor{red}{\textbf{52.9}}   & \textcolor{red}{\textbf{90.4}} & \textcolor{red}{\textbf{72.8}} & \textcolor{red}{\textbf{79.9}}   & \textcolor{red}{\textbf{85.0}}   & \textcolor{red}{\textbf{81.3 (+1.9)}}  \\
\bottomrule
\end{tabular}
}
\caption{\textbf{Performance of different models across 7 Natural Language Understanding (NLU) tasks.} \textbf{\textcolor{red}{Red}} results indicate our method brings improvement over the regular decoding, while \textbf{\textcolor{green}{Green}} denote no improvement.}
\label{tab:task_model}
\end{table*}

\begin{table}
\centering
\scalebox{0.85}{
\begin{tabular}{cccc} 
\toprule
\textbf{Model}                        & \textbf{Decoding} & \textbf{TREC}             & \textbf{Dbpedia}          \\ 
\midrule
\multirow{2}{*}{\textbf{Llama3.2-1B}} & \textbf{Regular}  & 40.0                                  & 85.6                                   \\
                                      & \textbf{Ours}     & \textbf{\textcolor{red}{46.2 (+6.2)}} & \textbf{\textcolor{red}{90.5 (+4.9)}}  \\ 
\midrule
\multirow{2}{*}{\textbf{Llama3.2-3B}} & \textbf{Regular}  & 44.4                                  & 83.1                                   \\
                                      & \textbf{Ours}     & \textbf{\textcolor{red}{49.6 (+5.2)}} & \textbf{\textcolor{red}{91.4 (+8.3)}}  \\ 
\midrule
\multirow{2}{*}{\textbf{Llama3.1-8B}} & \textbf{Regular}  & 41.0                                  & 87.5                                   \\
                                      & \textbf{Ours}     & \textbf{\textcolor{red}{46.6 (+5.6)}} & \textbf{\textcolor{red}{93.8 (6.3)}}   \\
\bottomrule
\end{tabular}}
\caption{\textbf{Average performance of two datasets with larger target classes.} \textbf{\textcolor{red}{Red}} results indicate that our method brings improvement over the regular decoding.}
\label{tab:mul_class}
\end{table}

\section{Analysis}
\label{subsect:analysis}
To further explore the impact of different factors on the effectiveness of our method, we conduct further analysis with the Llama3.2-8B models.

\paragraph{Effects of Different Negative In-context Examples.} As mentioned in Section \ref{subsec:cd}, the choice of negative in-context examples is important to the performance of our methods. Here, we conduct contrastive experiments to analyze the impact of different negative examples. Specifically, we refer to the selected negative examples as \textbf{Input}, if the input-label mapping is altered by modifying the inputs of the demonstrations. Additionally, we construct another variant, \textbf{Label}, in which the labels of the demonstrations are changed. For comparison, we also include \textbf{NULL}, which does not use any negative demonstrations, similar to ~\citet{shi-etal-2024-trusting}. The results in Table~\ref{tab:neg_exp} show that \textbf{Input} outperforms the other counterparts, thus leaving it as our default setting in this work.

\begin{table}
\centering
\scalebox{0.95}{
\begin{tabular}{cccc} 
\toprule
\multirow{2}{*}{\textbf{Method}} & \multicolumn{3}{c}{\textbf{Selection Method}}                                                                             \\ 
\cmidrule(lr){2-4}
                                 & \textbf{Random}                        & \textbf{BM25}                          & \textbf{TopK}                           \\ 
\midrule
\textbf{Regular Decoding}        & 77.6                                   & 79.7                                   & 80.2                                    \\ 
\midrule
\multicolumn{4}{l}{\textbf{\textit{Equipped with our method}}}                                                                                               \\ 
\hdashline
\textbf{+NULL}                   & \textbf{\textcolor{green}{73.0}}       & \textcolor{green}{\textbf{75.8}}       & \textcolor{green}{76.5}                 \\
\textbf{+Label}                  & \textbf{\textcolor{green}{77.3}}                                       & \textbf{\textcolor{green}{79.5}}                                       &  \textbf{\textcolor{green}{80.0}}                                       \\
\textbf{+Input}                  & \textcolor{red}{\textbf{79.4}} & \textcolor{red}{\textbf{80.8}} & \textcolor{red}{\textbf{80.9}}  \\
\bottomrule
\end{tabular}}
\caption{\textbf{Average performance with different negative in-context examples.} \textbf{\textcolor{red}{Red}} results indicate that our method brings improvement over the regular decoding, while \textbf{\textcolor{green}{Green}} results denote no improvement.}
\label{tab:neg_exp}
\end{table}

\paragraph{Differences between the positive and negative examples.} To verify whether our proposed method can truly lead to models to contrast the positive and negative in-context examples, we calculate the average KL divergence between the output distributions and report the results in Table~\ref{tab:kl}, we can notice that our method can get large KL divergence in most datasets, which means that the output distributions of positive and negative in-context examples are notably different. This demonstrates that our method can truly lead to models to contrast the positive and negative in-context examples.

\begin{table}
\centering
\scalebox{0.6}{
\begin{tabular}{cccccccc} 
\toprule
                        & \textbf{SST2} & \textbf{CR} & \textbf{SST5} & \textbf{Subj} & \textbf{QNLI} & \textbf{MNLI} & \textbf{AGNEWS}  \\ 
\midrule
\textbf{KL\_divergence} & 0.64          & 0.48        & 0.43          & 0.49          & 0.04          & 0.27          & 0.79             \\
\bottomrule
\end{tabular}}
\caption{\textbf{The average KL divergence} between the normalized output distributions with positive and negative in-context examples with Llama3.2-8B.}
\label{tab:kl}
\end{table}

\begin{figure}[t!]
    \centering
    \includegraphics[width=1\columnwidth]{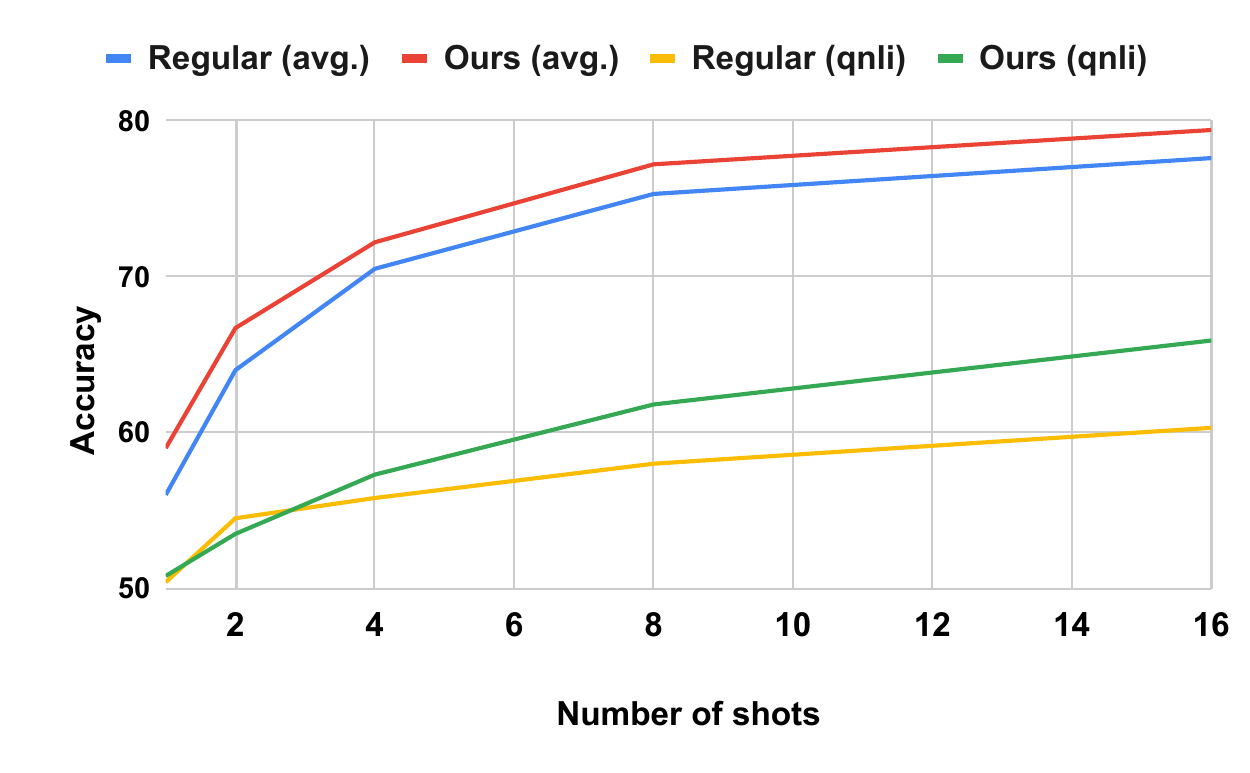}
    \caption{\textbf{The performance with different shots.}}
    \label{fig:shots}
\end{figure}

\paragraph{Effects of Different number of shots.}
We gradually increase the number of in-context examples (denoted as N) from 1 to 16 to verify the influence of the number of shots in our method. Figure~\ref{fig:shots} reports the average performance of 7 NLU tasks and the different task QNLI. We see that our method can consistently outperform the regular decoding method with a different number of shots on average. For the task QNLI, as the number of shots increases, the performance gains of our method also improve. We attribute this to the model acquiring more input-label mapping information from the demonstrations, which aligns with previous findings~\cite{pan-etal-2023-context}.

\paragraph{Effects of $\alpha$.} The factor $\alpha$ in Eq.~\ref{eq:log}, which controls the importance of input-label mapping information, is an important hyper-parameter. In this part, we analyze its influence by evaluating the performance on SST5 and MNLI varying $\alpha$ from 0 to 2. The results on Table~\ref{tab:alpha} show that: 1) the performance improves with the increase of $\alpha$, and it becomes stable when $\alpha \geq 1.0$, we set $\alpha=1$ as default; 2) For advanced demonstration selection methods(e.g.TopK), too large positive $\alpha$ values lead to performance degradation. 

\begin{table}
\centering
\scalebox{0.86}{
\begin{tabular}{ccccccc} 
\toprule
\multirow{2}{*}{\textbf{Dataset}} & \multirow{2}{*}{\textbf{Method}} & \multicolumn{5}{c}{\textbf{$\alpha$}}                                                                              \\ 
\cmidrule(lr){3-7}
                                  &                                  & 0.0  & 0.5  & 1.0                            & 1.5                            & 2.0                             \\ 
\midrule
\multirow{3}{*}{\textbf{SST5}}    & \textbf{Random}                  & 48.0 & 49.1 & 49.3                           & 49.3                           & \textcolor{red}{\textbf{49.5}}  \\
                                  & \textbf{BM25}                    & 53.0 & \textcolor{red}{\textbf{53.6}} & 53.3                           & 53.2 & 53.1                            \\
                                  & \textbf{TopK}                    & 53.0 & 53.2 & \textbf{\textcolor{red}{53.2}} & 52.9                           & 52.5                            \\ 
\midrule
\multirow{3}{*}{\textbf{MNLI}}    & \textbf{Random}                  & 65.3 & 66.8 & 67.5                           & 67.8                           & \textbf{\textcolor{red}{68.1}}  \\
                                  & \textbf{BM25}                    & 65.8 & 66.6 & 67.1                           & 67.4                           & \textbf{\textcolor{red}{67.6}}  \\
                                  & \textbf{TopK}                    & 65.9 & 67.0 & 67.4 & \textbf{\textcolor{red}{67.7}}                           & 67.7                            \\
\bottomrule
\end{tabular}
}
\caption{\textbf{The SST5 and MNLI performance with different $\alpha$.}}
\label{tab:alpha}
\end{table}

\section{Conclusion}
Large language models suffer from insufficient attention to the input-label mapping compared to their prior knowledge in in-context learning, leading to an unfaithful generation of the input query. In this work, we present a simple yet effective in-context contrastive decoding method that highlights input-label mapping by contrasting positive and negative in-context examples. Our experiments across various datasets and model architectures demonstrate the effectiveness and broad applicability of our approach, confirming its potential to enhance in-context learning.

\section*{Limitations}
While the results presented in this paper demonstrate the effectiveness of our In-Context Contrastive Decoding (ICCD) method, there are a few limitations that warrant future exploration. First, our experiments were conducted on models up to 8B parameters, primarily due to computational limitations. Extending our method to even larger models (e.g., 70B parameters) could provide further insights into its scalability and effectiveness. Second, while our method shows promise across various Natural Language Understanding (NLU) tasks, its performance in specialized domains, such as legal or medical texts, has yet to be thoroughly examined. Third, our method requires additional forward passes to compute contrastive distributions. Although these passes are executed in parallel, the overall inference time may still increase. Future work will explore the generalizability of ICCD to these domains, as well as investigate its interaction with domain-specific datasets. Additionally, while we focused on classification tasks, other NLP tasks like text generation, machine translation, and summarization remain unexplored.

\section*{Acknowledges}
We are grateful to the anonymous reviewers and the area chair for their insightful comments and suggestions. This work is supported by the National Natural Science Foundation of China (No. 62377002). This project is supported by the National Research Foundation, Singapore, under its NRF Professorship Award No. NRF-P2024-001.
\bibliography{arxiv}

\newpage
\appendix

\section{Datasets}
Natural Language Understanding (NLU) Dataset information is detailed in Table~\ref{tab:dataset}. All NLU datasets are loaded from the HuggingFace Hub. For most NLU datasets, we report the results on the test set; while for the datasets MNLI and QNLI, we report the results on the validation set due to restricted access to their test sets. 

\begin{table*}[t]
    \centering
    \resizebox{0.8\linewidth}{!}{
    \begin{tabular}{lcccc}
    \toprule
    \textbf{Dataset} & \textbf{Task} & \textbf{\# of Classes} & \textbf{Data Split} \\ \midrule
     \textbf{SST-2}  &  Sentiment Classification & 2 & 6920/872/1821 \\
     \textbf{SST-5}  &  Sentiment Classification & 5 & 8544/1101/2210 \\
     \textbf{CR}  &  Sentiment Classification & 2 & 3394/0/376 \\
     \textbf{Subj}  &  Subjectivity Analysis & 2 & 8000/0/2000 \\
     \textbf{AgNews} &  Topic Classification & 4 & 120000/0/7600 \\
     \textbf{MNLI}   &  Natural Language Inference & 3 & 392702/19647/19643 \\
      \textbf{QNLI}  &  Natural Language Inference & 2 & 104743/5463/5463 \\
      \bottomrule
      \end{tabular}}
    \caption{\textbf{Details of NLU datasets.}}
    \label{tab:dataset}
\end{table*}

\section{Templates}
\begin{table*}[t ]
\centering
\resizebox{0.8\linewidth}{!}{
\begin{tabular}{lll}
\toprule
\textbf{Task} & \textbf{Prompt} & \textbf{Class} \\
\hline
\multirow{2}{*}{SST-2} & Review: "<X>" Sentiment: positive & positive \\ 
& Review: "<X>" Sentiment: negative & negative \\
\midrule
\multirow{5}{*}{SST-5}
& Review: "<X>" Sentiment: terrible  & terrible \\ 
& Review: "<X>" Sentiment: bad  & bad \\ 
& Review: "<X>" Sentiment: okay  & okay \\ 
& Review: "<X>" Sentiment: good  & good \\ 
& Review: "<X>" Sentiment: great  & great \\ 
\midrule
\multirow{2}{*}{Subj}
& Input: "<X>" Type: objective  & objective \\ 
& Input: "<X>" Type: subjective  & subjective \\ 
\midrule
\multirow{2}{*}{CR} & Review: "<X>" Sentiment: positive & positive \\ 
& Review: "<X>" Sentiment: negative & negative \\
\midrule
\multirow{4}{*}{AgNews}
& Input: "<X>" Type: world  & World \\ 
& Input: "<X>" Type: sports  & Sports \\ 
& Input: "<X>" Type: business  & Business \\ 
& Input: "<X>" Type: technology  & Sci/Tech \\ 
\midrule
\multirow{3}{*}{MNLI}
& Premise: <C> Hypothesis: <X> Prediction: entailment  & Entailment \\ 
& Premise: <C> Hypothesis: <X> Prediction: neutral  & Neutral \\ 
& Premise: <C> Hypothesis: <X>? Prediction: contradiction  & Contradiction \\ 
\midrule
\multirow{2}{*}{QNLI}
& <C> Can we know <X>? Yes.  & Entailment \\ 
& <C> Can we know <X>? No.  & Contradiction \\ 
\midrule
\end{tabular}
}
\caption{\textbf{Templates of NLU tasks.} Placeholders (e.g., <X> and <C>) will be replaced by real inputs.}
\label{tab:templates}
\end{table*}

The templates of NLU tasks used in this paper are detailed in Table~\ref{tab:templates}.

\end{document}